\documentclass[11pt,a4paper]{article}
\usepackage[hyperref]{emnlp2020}
\usepackage{times}
\usepackage{latexsym}

\usepackage{microtype}

\aclfinalcopy % Uncomment this line for the final submission
\usepackage{amsmath,amsfonts,bm}

\def\eqref#1{equation~\ref{#1}}
\def\1{\bm{1}}

\def\va{{\bm{a}}}

\def\ve{{\bm{e}}}

\def\vg{{\bm{g}}}
\def\vh{{\bm{h}}}

\def\vs{{\bm{s}}}

\def\vv{{\bm{v}}}

\def\vx{{\bm{x}}}
\def\vy{{\bm{y}}}
\def\vz{{\bm{z}}}

\def\mW{{\bm{W}}}

\DeclareMathAlphabet{\mathsfit}{\encodingdefault}{\sfdefault}{m}{sl}
\SetMathAlphabet{\mathsfit}{bold}{\encodingdefault}{\sfdefault}{bx}{n}

\newcommand{\R}{\mathbb{R}}

\newcommand{\SG}{\mathrm{SG}}

\usepackage{hyperref}
\usepackage{url}
\usepackage{graphicx}
\usepackage[font=small]{caption}
\usepackage{booktabs}
\usepackage{tabulary}

\usepackage{multirow}
\usepackage{mathtools}
\mathtoolsset{showonlyrefs=true}

\definecolor{forestgreen}{HTML}{009B55}
\definecolor{sepia}{HTML}{671800}
\definecolor{midnightblue}{HTML}{006795}
\definecolor{orangered}{HTML}{E24C00}

\title{Sticking to the Facts: Confident Decoding for Faithful Data-to-Text Generation}

\author{Ran Tian, Shashi Narayan, Thibault Sellam \& Ankur P.~Parikh \\
  Google Research \\
  \texttt{\{tianran,shashinarayan,tsellam,aparikh\}@google.com}
}

\begin{document}
\maketitle

\begin{abstract}
We address the issue of \emph{hallucination} in data-to-text generation, i.e., reducing the generation of text that is unsupported by the source. We conjecture that hallucination can be caused by an encoder-decoder model generating content phrases without attending to the source; so we propose a confidence score to ensure that the model attends to the source whenever necessary, as well as a variational Bayes training framework that can learn the score from data.
Experiments on the WikiBio~\citep{lebret2016neural} dataset show that our approach is more faithful to the source than existing state-of-the-art approaches, according to both PARENT score~\citep{dhingra2019handling} and human evaluation. We also report strong results on the WebNLG~\citep{gardent-etal-2017-creating} dataset.
\end{abstract}

\section{Introduction}
The task of generating natural language text $\vy$ from a source content $\vx$ is the essence of many NLP applications, such as summarization~\citep{mani1999advances}, machine translation~\citep{koehn2009statistical}, and data-to-text generation~\citep{kukich1983design,mckeown1992text}. While traditionally done with template-based approaches~\citep{becker2002practical,foster2004techniques,gatt2009simplenlg,reiter2005choosing}, recent neural encoder-decoder models~\citep{kalchbrenner2013recurrent,sutskever2014sequence,cho2014learning,bahdanau2014neural} have demonstrated remarkable ability to generate fluent text without cumbersome handcrafted rules and templates~\citep{rush2015neural,radford2019language}.

However, encoder-decoder models have been shown to be prone to \textit{hallucination}, i.e., generating text that is fluent but \emph{unfaithful} to the source~\citep{vinyals2015neural,koehn2017six,wiseman2017challenges,maynez2020faithfulness}. This severe shortcoming can often limit the use of neural approaches in many real world systems, where it is not acceptable to produce output that is even occasionally unfaithful.

\begin{figure}[t!]
\centering
\footnotesize
{\renewcommand{\arraystretch}{1.4}
\begin{tabular}{l p{5cm}}
    \toprule
    \multicolumn{2}{c}{\textbf{Wikipedia Infobox}} \\ \hline
    \multicolumn{2}{c}{\textbf{Frank Lino}} \\
    \textbf{Caption} &  FBI surveillance                                photo \\
    \textbf{Birth date} & October 30, 1938 \\
    \textbf{Birth place} & Gravesend, Brooklyn, New York, United States \\
    \midrule
    \multicolumn{2}{p{7cm}}{\textbf{Reference:} \textit{Frank ``Curly'' Lino (born October 30, 1938 Brooklyn)  is a Sicilian-American Caporegime in the Bonanno crime family who later became an informant.}} \\
    \multicolumn{2}{p{7cm}}{\textbf{Baseline:} \textit{Frank Lino (born October 30, 1938 in Brooklyn, New York, United States) is an American \textcolor{orangered}{criminal defense attorney}.}} \\
    \multicolumn{2}{p{7cm}}{\textbf{Our model:} \textit{Frank Lino (born October 30, 1938 in Brooklyn, New York, United States) is an American.}} \\
    \bottomrule
\end{tabular}}
\caption{Example in the WikiBio dataset~\citep{lebret2016neural} showing the biography of \emph{Frank Lino}. The baseline Pointer-Generator~\citep{see2017get} exhibits hallucination.}
\label{fig:hallucination-ex}
% \vspace{-0.2cm}
\end{figure}

In this work, we address the issue of hallucination in data-to-text generation, where the source content $\vx$ is a structured table and the text $\vy$ is a description of the table -- a relatively easy setting to objectively evaluate the faithfulness of generation. We show an example from the WikiBio dataset~\citep{lebret2016neural} in Figure~\ref{fig:hallucination-ex}; the task is to generate a sentence summarizing a tabular biography of a person. The output of a strong generation baseline, the Pointer-Generator~\citep{see2017get}, contains a phrase \emph{criminal defense attorney} that is incorrect and cannot be supported by the infobox table  (but loosely related to \emph{FBI} in the table). Note that the reference also contains information such as \emph{bonanno crime family} and \emph{informant} that are true, but cannot be inferred from the infobox; this \emph{source-reference divergence} exists in many large-scale generation datasets~\citep{wiseman2017challenges}, and might encourage a generation model to output phrases that are unsupported by the source.

However, the issue is not limited to divergence between source and reference since hallucination can appear even with cleaned references~\citep{parikh2020totto}. Rather, the underlying problem is that the model learns wrong correlations between different parts of the training data. As the data and models get larger and more complicated, learning wrong correlations might always become an issue because of abundant inter-related factors in play. Thus, we need methodology to control neural networks and pose our prior knowledge on ``correct correlations'' to the models. This is an important, yet less addressed challenge in deep learning.

In this work, we pose a ``confidence prior'' to encoder-decoder models, by carefully reconsidering the two components in a decoder: attention to the source and language modeling. Our prior knowledge is that a model should attend to the source when generating a word, as long as the word conveys source information. Wrongly associating a content phrase (e.g.~\emph{defense attorney}) to the language model, simply because it seems more fluent (e.g.~\emph{criminal defense attorney} is fluent), might be a major cause of hallucination (\S~\ref{sec:srcsens}).

Therefore, we design a confidence score to detect hallucination, by using an attention score to measure how much the model is attending to the source, and a language model to judge if a word conveys source information (\S~\ref{sec:confscore}). Then, we propose a variational Bayes training framework that can ensure a model to generate with high confidence, while learning the confidence score parameters at the same time (\S~\ref{sec:variational}). Experiments on the WikiBio dataset demonstrate that our approach is considerably more faithful to the source than existing state-of-the-art solutions, according to both PARENT score~\citep{dhingra2019handling} and human evaluation (\S~\ref{sec:wikibiores}). We also report strong results on the WebNLG~\citep{gardent-etal-2017-creating} dataset (\S~\ref{sec:webnlgres}).

\section{Preliminaries}

We first review the existing encoder-decoder model~\citep{bahdanau2014neural} which this work is based on.
%with one stop-gradient ($\text{SG}$) tweak.
Let $\vx = x_1 x_2\dots x_S$, be the source input of length $S$ and $\vy = y_1 y_2\dots y_T$ be the target sequence of length $T$. Each token $x_i, y_i$ takes one value from a vocabulary $V$. The goal is to model the conditional distribution $P(\vy | \vx)=\prod_{t=1}^{T} P(y_t | \vy_{<t}, \vx)$,
% \begin{align}
% \label{eq:seq2seq}
% P\bigl(\vy \mid \vx\bigr) = \prod_{t=1}^{T} P\bigl(y_t \mid \vy_{<t}, \vx\bigr),
% \end{align}
where $\vy_{<t}=y_1\dots y_{t-1}$ is the prefix of $\vy$ up to the $(t-1)^\text{th}$ token. The source can be encoded by any neural network function \textbf{enc}, such as a convolutional neural network~\citep[CNN,][]{lecun1990handwritten}, long-short-term memory~\citep[LSTM,][]{hochreiter1997long}, or Transformer~\citep{vaswani2017attention}. Let $\vs_1,...,\vs_{S} = \textbf{enc}(x_1,...,x_S)$. Define $\ve_{x} \in \R^d$ as the $d$ dimensional embedding of token $x$. Then, the probability of each target token is computed as:
\begin{equation}
P\left(y_t \mid \vy_{<t}, \vx\right)
= \frac{\exp{ (\vv_{t}^\top\ve_{y_{t}}})}{\sum_{y\in V} (\exp{ \vv_{t}^\top\ve_{y}})} \label{eq:softmax}
\end{equation}
where the context vector $\vv_{t}$ is given by:
\begin{equation}
\vv_{t}=\va_{t}+\vh_{t}=\sum_{s=1}^S \alpha_{s,t} \vs_{s}+\vh_{t} \label{eq:context_vector}
\end{equation}
Here, $\alpha_{s,t}$ is an attention weight of the prefix $\vy_{<t}$ attending to source position $s$, for which we use bilinear attention~\citep[Eq.~\eqref{eq:attention},][]{DBLP:conf/emnlp/LuongPM15}; and $\vh_{t}$ is given by an RNN\footnote{While it is possible our approach could extend to other types of decoders, our current formulation of the confidence score specifically uses RNN with attention.} (Eq.~\eqref{eq:rnn}, where $[\cdot]$ denotes concatenation):
\begin{align}
\alpha_{s,t} & = \frac{\exp(\vs_s^{\top}\mW\vh_{t})}{\sum_{s'} \exp(\vs_{s'}^{\top}\mW\vh_{t})}
\label{eq:attention}\\
\vh_{t} & =\text{RNN}(\vh_{t-1}, [\ve_{y_{t-1}}, \va_{t-1}])
\label{eq:rnn}
\end{align}

In case the encoder-decoder is equipped with a copy mechanism, the generation probability is mixed with a probability of copying from the source \citep{gu2016incorporating,see2017get}:
\begin{multline}
\tilde{P}(y_t| \vy_{<t}, \vx) = p^{\text{gen}}_{t} P(y_t| \vy_{<t}, \vx) \\
 + (1-p^{\text{gen}}_{t}) \sum_{s:x_{s}=y_{t}}\beta_{s,t}
\end{multline}
where $p^{\text{gen}}_{t}$ is the probability of doing generation instead of copying at step $t$, and $\beta_{s,t}$ is an attention weight that the copy mechanism is paying to position $s$ in the source. The sum is taken over all positions $s$ where the word $x_{s}$ is the same as $y_{t}$.

\section{Modeling Confident Decoding}
%\label{sec:modeling}

\begin{figure*}[t!]
\centering
\includegraphics[scale=0.3,viewport=0 0 52cm 10cm,clip]{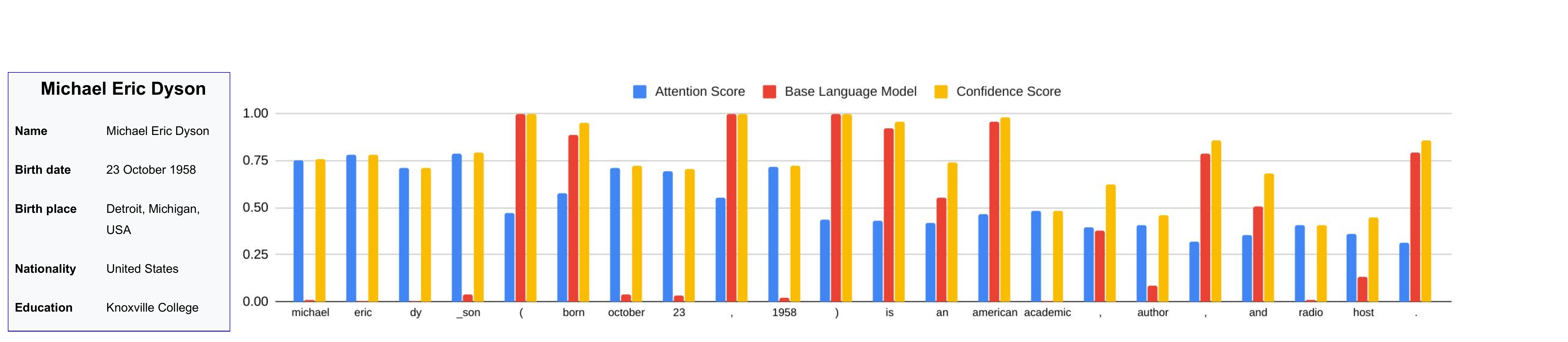}
\caption{Example of our learned attention score, base language model probability, and confidence score.}
\label{fig:scores}
\vspace{-0.2cm}
\end{figure*}

%In this section, we mathematically describe our approach. We first define the confidence score (\S~\ref{sec:confscore}), then derive a training objective (\S~\ref{sec:variational}) that jointly learns and the confidence score by maximizing the data likelihood. Finally we discuss two additional techniques to utilize the confidence score for decoding (\S~\ref{sec:calibration}).

%\subsection{Confidence Score}
\label{sec:confscore}
In this section, we mathematically describe our approach.
For each decoder position $t$, we define the following \textbf{confidence score} $C_{t}(y_t)$ to detect hallucination:
\begin{equation}
C_{t}(y_t):= A_{t} + (1-A_{t})P_{B}\bigl(y_t \mid \vy_{<t}\bigr) \label{eq:confidence_score}
\end{equation}
 Here, $A_{t} \in [0,1]$ is the attention score (see below ``Attention Score'') which indicates how much the model is generating based on the source; e.g., it should be close to $1$ for content words copied from the source. $P_{B}(y_t | \vy_{<t})$ is the probability of a tailored language model (see below ``Base Language Model''), that should be high for templatic words but low for words that convey source information.
 
 Figure~\ref{fig:scores} shows an example. Templatic words (e.g.~\emph{born}, \emph{is}) do not need support from the source, and they have high probability by the base language model and high confidence score, regardless of the attention score. On the other hand, tokens that convey source information (e.g.~\emph{Michael}, \emph{author}) have low probability by the base language model, and their confidence scores depend on the attention scores. A low confidence score indicates that a word conveying source information is generated by the model without paying attention to the source; which, we conjecture, is a signal of hallucination. Indeed, in Figure~\ref{fig:scores}, the tokens with lower confidence scores (e.g.~\emph{author}, \emph{radio host}) are not supported by the table. This example is taken from the WikiBio validation set (cf.~Appendix for more).

% So we have $C_{t}(y_t)\in [0,1]$ where we conjecture a low confidence score indicates hallucination. Intuitively, for function words and template elements, we expect $P_{B}(y_t | \vy_{<t})$ to be high, so the confidence score will be high no matter what the attention score is. Conversely, $\text{RNN}_{B}$ should be less predictive for tokens that convey source information, so $P_{B}(y_t | \vy_{<t})$ will be low and the confidence score which depends on the attention score in this case, can detect hallucination.
 
  %Now, we give more details on how the attention score and the base language model are defined.

\paragraph{Attention Score} The attention score should measure how much a token is generated based on source. We modify the conventional attention mechanism (Eq.~\eqref{eq:attention} \eqref{eq:rnn}) in two ways to make such measurement easier. First, we make the attention weights sum to less than $1$, so that the model can choose ``not to attend''; this is achieved by adding a constant $1$ to the denominator of Eq.~\eqref{eq:attention2}. Second, instead of concatenating the previous attention vector to the input of RNN, we only use $\va_{t-1}$ in calculation of the current attention weights, so that \emph{the hidden states of the RNN no longer contain any source information} (Eq.~\eqref{eq:attention2}~\eqref{eq:rnn2}):
\begin{align}
\alpha_{s,t} & = \frac{\exp(\vs_s^{\top}\mW[\vh_{t},\va_{t-1}])}{1+\sum_{s'} \exp(\vs_{s'}^{\top}\mW[\vh_{t},\va_{t-1}])}
\label{eq:attention2}\\
\vh_{t} & =\text{RNN}(\vh_{t-1}, \ve_{y_{t-1}})
\label{eq:rnn2}
\end{align}
Then, because the next token is generated by the context vector $\vv_{t}=\va_{t}+\vh_{t}$ in Eq.~\eqref{eq:context_vector}, and all the source information in $\vv_{t}$ comes from $\va_{t}$, we define the \textbf{attention score} $A_{t}$ as below to measure how much $\va_{t}$ affects $\vv_{t}$ (where $\lVert\cdot\rVert$ denotes Euclidean norm):
\begin{equation}
\label{eq:attention_score}
A_{t}:=\frac{\lVert\va_{t}\rVert}{\frac{1}{2}\bigl(\lVert\va_{t}\rVert+\lVert\vh_{t}\rVert+\lVert\vv_{t}\rVert\bigr)}
\end{equation}
We have $A_{t} \in [0,1]$ by triangle inequality. In an extreme case, when $\vh_{t}$ is completely cancelled out by $\va_{t}$ in the sum $\vv_{t}$, the attention score equals $1$. When the model has a copy mechanism, we can refine $A_{t}$ with the copying probability:
\begin{equation}
\tilde{A}_{t}:=p^{\text{gen}}_{t}A_{t}+(1-p^{\text{gen}}_{t})
\end{equation}

In practice, we have confirmed that our modification of the attention mechanism (Eq.~\eqref{eq:attention2}~\eqref{eq:rnn2}) does not impact the quality of data-to-text generation (\S~\ref{sec:webnlgres}); the observed attention score has a reasonable range of $0.6\sim 0.9$ for tokens supported by the source, and $0.2\sim 0.5$ for templatic words or hallucinated tokens (Figure~\ref{fig:scores}).

\begin{figure*}[t!]
\centering
\includegraphics[scale=0.38,viewport=0 0 38cm 10cm,clip]{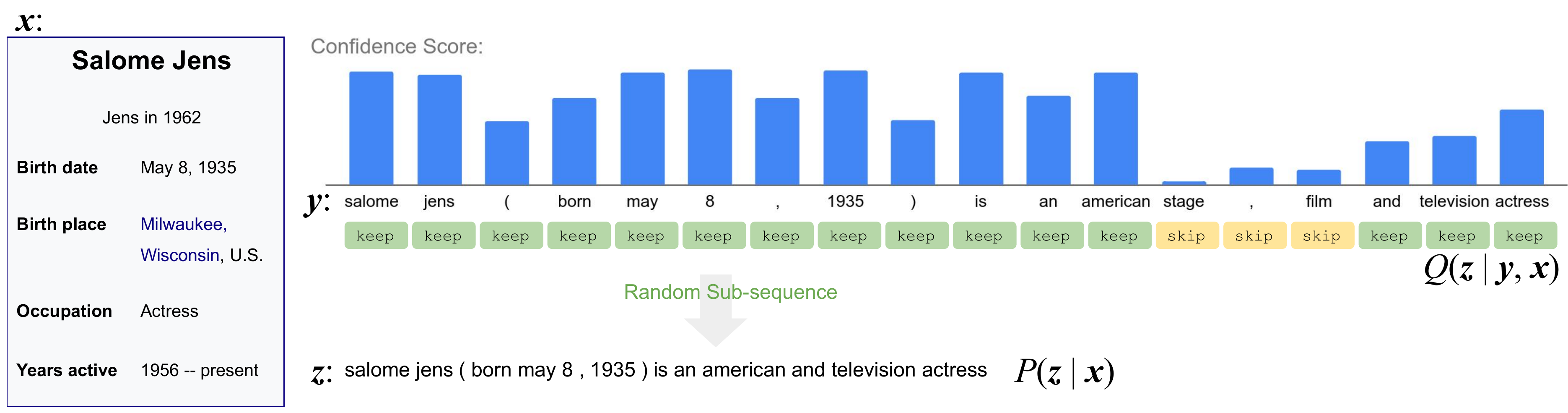}
\caption{Example of sampling a sub-sequence according to the confidence score. Our variational Bayes objective combines the sampling probability $Q(\vz |\vy, \vx)$ and the generation probability $P(\vz |\vx)$.}
\label{fig:subsequence}
\vspace{-0.3cm}
\end{figure*}

\paragraph{Base Language Model} In order to be effective, $\text{RNN}_{B}$ should be able to learn ``soft templates'' rather than simply fluent text. Unfortunately using an ordinary unconditioned language model for $P_B(y_t | \vy_{<t})$ can be problematic, since the model can learn source-specific knowledge through $\vy_{<t}$. For instance if there is only one person named \emph{Walter} in the training data, and he is a pilot, then the language model might learn \emph{Walter is a pilot} as a fixed generation pattern. We tailor $\text{RNN}_{B}$ to reduce such artifacts by down-weighting input embeddings that are associated with high attention scores (and thus are source-specific):
\begin{align}
\vg_{t}&=\text{RNN}_{B}\bigl(\vg_{t-1},
(1-w)\ve_{y_{t-1}}+w\ve_{\text{\textless{src}\textgreater}}\bigr)\\
w&:=\SG(A_{t-1})
\end{align}
 $\ve_{\text{\textless{src}\textgreater}}$ is a special trainable embedding and $\SG$ denotes stop-gradient, which is necessary to prevent the training of $\text{RNN}_{B}$ from affecting the encoder-decoder model through the attention score\footnote{In our preliminary experiments, omitting $\SG$ here and training $\text{RNN}_{B}$ jointly with the encoder-decoder will result in most attention scores becoming less than $0.5$.}. Practically, we have confirmed that $\text{RNN}_{B}$ learns soft generation templates (Figure~\ref{fig:scores}) and describe how it is trained in \S~\ref{sec:variational}.

\subsection{Training with Confident Sub-sequence Sampling}
\label{sec:variational}

How do we train a model to generate confidently, using the confidence score we just proposed? Note that the confidence score itself has trainable parameters (i.e., attention score and the parameters of $\text{RNN}_{B}$). Our idea is to assume a latent ``confident sub-sequence'' of the target for each training example, and learn the latent sub-sequence by sampling according to the confidence score. As training progresses, the confidence score improves and the sampled sub-sequences contain only the parts of the target that are faithful to the source. An example is given in Figure~\ref{fig:subsequence} where  the model learns to assign low scores to the tokens \textit{stage, film} so the sampled subsequence only contains information faithful to the source.

Formally, for each target $\vy = y_1 y_2\dots y_T$, we define $\vz = z_1 z_2\dots z_R = y_{\iota(1)} y_{\iota(2)}\dots y_{\iota(R)}$ as a latent sub-sequence of $\vy$, which consists of confident tokens of length $R$. Here, $\iota :|R|\rightarrow |T|$ is an inclusion of indices. We regard $\vz$ as a sequential ``\texttt{keep} or \texttt{skip}'' labeling over $\vy$, and sample $\vz$ from the probability distribution $Q(\vz |\vy, \vx)=\prod_{t=1}^{T}Q_{t}$:
\begin{equation}
%Q\bigl(\vz \mid\vy, \vx\bigr)=\prod_{t=1}^{T}Q_{t}\quad\quad\quad\quad\quad\quad\\
Q_{t}=
\begin{cases}
Q_{t}(\texttt{keep})=\dfrac{C_{t}(y_t)^\rho}{C_{t}(y_t)^\rho+\gamma} & \text{If $y_{t}\in\vz$} \\
Q_{t}(\texttt{skip})=1-Q_{t}(\texttt{keep}) & \text{If $y_{t}\notin\vz$}
\end{cases}
\end{equation}
Here, $\rho$ and $\gamma$ are trainable parameters initialized to $0$ and $1$, respectively. As $\rho$ gets larger, it more strictly enforces our prior knowledge of faithful generation: Every (kept) token should have a high confidence score, which means either the encoder-decoder is paying attention to the source, or the token is a template element that does not convey source information. Empirically, the trained $\rho$ in our model indeed converges to a positive value (e.g.~about $3.4$ on the WikiBio dataset).

For each training example $(\vx, \vy)$, our objective is to minimize the following generation cost:
\begin{align}
\label{eq:bayes_rule}
\mathcal{L}(\vy, \vx) &= - \log P(\vy | \vx) \\
&= -\log \frac{P(\vy |\vz, \vx)\,P(\vz |\vx)}{P(\vz |\vy, \vx)}
\end{align}
Above we have applied the Bayes rule; so we model
$P(\vy |\vz, \vx)$ and $P(\vz |\vx)$, instead of $P(\vy | \vx)$. We set $P(\vz | \vx)$ to be the encoder-decoder model as described before. Since $P(\vy |\vz, \vx)$ is not used in test, we simply assume it remembers all the training examples and, when given an $\vx$ that appears in training data, gives a probability $1$ to the gold reference $\vy$ and $0$ to all others. Hence, we can set $P(\vy |\vz, \vx)=1$ here, and our modeling efforts focus on $P(\vz | \vx)$.
%We assume that for each training example, the gold reference $\vy$ is fixed and can be uniquely recovered from its confident sub-sequence $\vz$, for simplicity. Hence $P(\vy |\vz, \vx)=1$.\footnote{To put it in another way: $P(\vy |\vz, \vx)$ is a modeling factor up to our choice. Since we want to treat all sub-sequences of $\vy$ equally, $P(\vy |\vz, \vx)$ does not depend on $\vz$.}

\paragraph{Variational Bayes} Unfortunately, the posterior $P(\vz |\vy, \vx)$ in the above objective cannot be arbitrarily modeled\footnote{For one thing, the right hand side of Eq.~\eqref{eq:bayes_rule} should give the same value for any $\vz$, because the left hand side does not depend on $\vz$. This is a non-trivial constraint for $P(\vz |\vy, \vx)$.}.
%depends on the data and is unknown.
We thus employ a Variational Bayes scheme~\citep{koller2009probabilistic} and use our sampling probability $Q=Q(\vz |\vy, \vx)$ to approximate $P(\vz |\vy, \vx)$. By adding $\log Q$, we get
\begin{multline}
-\log P\bigl(\vy | \vx\bigr) = - \log\frac{Q}{P(\vz |\vy, \vx)} \\ + \log Q - \log P\bigl( \vz | \vx \bigr)
\end{multline}
Then, taking the expectation $\mathbb{E}_{Q}[\cdot]$ of both sides and because $\mathbb{E}_{Q}[\log\frac{Q}{P(\vz |\vy, \vx)}]=KL[Q || P(\vz |\vy, \vx)]\geq 0$, we get
\begin{equation}
-\log P\bigl(\vy | \vx\bigr) \leq \mathbb{E}_{Q}\Bigl[\log Q - \log P\bigl( \vz | \vx \bigr)\Bigr] \label{eq:variational_bayes}
\end{equation}
The variational Bayes objective is to minimize the upper bound on the right hand side of Eq.~\eqref{eq:variational_bayes}.

Thus, we have obtained a Bayesian model which implements our prior knowledge on confident generation, and the latent variables (i.e.~the confidence score) can be learned from data. Intuitively, the term $\mathbb{E}_{Q}\bigl[-\log P(\vz |\vx)\bigr]$ in Eq.~\eqref{eq:variational_bayes} ensures that the encoder-decoder only trains on sampled confident tokens, while the entropy term $\mathbb{E}_{Q}[\log Q]$ tries to increase the confidence score for tokens labeled \texttt{skip}. Eventually, the confidence scores converge so that the sub-sequences with higher log-likelihood (i.e.~lower perplexity) remain confident. Our experiments suggest that such sub-sequences are also fluent (\S~\ref{sec:wikibiores}), even if we did not explicitly require fluency in our sub-sequence sampling.

Importantly,  the base language model $\text{RNN}_{B}$ is trained in two ways: through $Q$ in the variational Bayes Objective and also by minimizing an additional $-\log P_B(\vz)$ term. Jointly training $\text{RNN}_{B}$ on the confident sub-sequence $\vz$ implicitly biases it toward more confident generation patterns.

In practice, it is also computationally expensive to explicitly calculate $\mathbb{E}_{Q}[\cdot]$ by enumerating all sub-sequences of $\vy$, because the number of sub-sequences is exponential to the length $T$. Thus, we apply a Monte Carlo method and calculate $\mathbb{E}_{Q}[\cdot]$ by sampling from $Q$. The overall loss is given by:
\begin{align}
\mathcal{L}(\vy, \vx)&:=\begin{multlined}[t]\frac{1}{K}\sum_{\substack{k=1\\ \vz_k\sim Q}}^{K}\bigl(H-\log P_B(\vz_k)\\+\SG(H)\log Q(\vz_k |\vy,\vx)\bigr)\end{multlined}\label{eq:loss}\\
H&:=\log Q(\vz_k|\vy,\vx) - \log P(\vz_k|\vx)
\end{align}
Here, the term $-\log P_B(\vz_k)$ comes from jointly training $\text{RNN}_{B}$, and $\SG(H)\log Q(\vz_k |\vy,\vx)$ is added to back-propagate gradients through the expectation  $\mathbb{E}_{Q}[\cdot]$~\citep{DBLP:conf/icml/PaisleyBJ12}.

\subsection{Calibration and \textless{null}\textgreater ~Token}
\label{sec:calibration}

Finally, we discuss two additional techniques to utilize the confidence score at inference time.

\paragraph{Calibration} With a model trained to generate confidently, one might still want to explicitly re-rank the generation probability at inference time toward more confident tokens. The \textbf{calibration} technique~\citep{braverman2019calibration} provides a way to learn such explicit re-ranking. It parameterizes a family of probability distributions that augment $P(y_t | \vy_{<t}, \vx)$ with some quantity that one cares, which in our case is the confidence score $C_t(y_t)$:
\begin{equation}
\hat{P}^{\kappa}(y_t| \vy_{<t}, \vx)\propto
\text{SG}(P(y_t| \vy_{<t}, \vx))\text{SG}(C_t(y_t))^{\kappa} %{\sum_{w\in V}\text{SG}(C(w | \vy_{<t}, \vx))^{\kappa}\text{SG}(P(w| \vy_{<t}, \vx))}.
\end{equation}
Here, $\kappa$ is a trainable parameter, and the right hand side is normalized so that $\hat{P}^{\kappa}(y_t| \vy_{<t}, \vx)$ sums to $1$. For training, we add another $-\log\hat{P}^{\kappa}(\vz_k|\vx)$ term into Eq.~\eqref{eq:loss}; note that $\SG$ (stop-gradient) prevents $P(y_t | \vy_{<t}, \vx)$ and $C_t(y_t)$ from being affected; only $\kappa$ is trained by this term. Since the family $\hat{P}^{\kappa}(\vy|\vx)$ has $P(\vy|\vx)$ as a special case (i.e.~$\kappa=0$), the optimized training perplexity of $\hat{P}^{\kappa}(\vy|\vx)$ will be at most $P(\vy|\vx)$. In practice, $\kappa$ is initialized to $0$ and found converging to a positive value (e.g.~about $0.65$ on WikiBio); so the calibration trick actually leads to re-ranking toward more confident tokens, without sacrificing training perplexity.

\paragraph{\textless{null}\textgreater ~token} If a token is generated with a confidence score lower than a certain threshold, we replace it with a special \textbf{\textless{null}\textgreater ~token}; the token is fed to the next step, and consecutive \textless{null}\textgreater ~tokens are shut out from the beam search. After beam search, all \textless{null}\textgreater{s} are deleted from the output sequence.
We slightly modify the sub-sequence sampling during training to be compatible with this strategy: Once a target token is labeled \texttt{skip}, it is replaced by a \textless{null}\textgreater ~instead of being skipped; only consecutive tokens labeled \texttt{skip} are actually skipped (i.e.~not being counted by the sampled sub-sequence $\vz$).
%~and fed to the next step; when consecutive tokens are skipped, the decoder RNN states are not updated and the consecutive tokens are not counted by the $-\log P(\vz |\vx)$ term in the variational Bayes objective (but are still counted by $Q$).
Intuitively, the \textless{null}\textgreater ~token mimics a ``pause and rethink'' strategy, making the generation process more robust against unconfident tokens. Empirically, we found that \textless{null}\textgreater ~token combined with length penalty~\citep{wu2016google} can drastically increase recall while maintaining precision in data-to-text generation (\S~\ref{sec:wikibiores}).

\section{Experiments}
\label{sec:exp}
We evaluate on the WikiBio~\citep{lebret2016neural} and WebNLG~\citep{gardent-etal-2017-creating} datasets. These datasets exhibit different levels of source-reference divergence and thus test our model in different regimes. Specifically, WikiBio is heuristically collected and 62\% of examples exhibit divergence~\citep{dhingra2019handling}, whereas WebNLG has human generated responses with less divergence.

WikiBio contains 728,321 infoboxes paired with biographies, taken from the Sep.-2015 dump of English Wikipedia, and split into train/valid/test sets in a 8:1:1 ratio. The biography text is the first sentence of the Wikipedia page ($26.1$ words on average). Infoboxes have $12.1$ non-empty fields on average. The WebNLG release v2.1 with constrained split~\citep{shimorina2018handling} contains 16,095 data inputs in the format of RDF triples, and 42,873 data-text pairs (i.e.~multiple references for each data input), splitted in a 8:1:1 ratio. The constrained split ensures that no RDF triple in the test set is in the train or dev set.

As a typical setting, we treat the data-to-text tasks as seq-to-seq prediction; infoboxes and RDF triples are linearized, with ``key/value''s or ``subject/relation/object''s separated by special tokens.

\subsection{Results on WikiBio}
\label{sec:wikibiores}

\begin{table*}[t!]
\footnotesize
	\centering
	\begin{tabulary}{0.5\textwidth}{@{}l | c@{\hskip 0.6cm} c@{\hskip 0.4cm} c | c@{\hskip 0.5cm} c@{\hskip 0.5cm} c@{}}
		\toprule
	    \multirow{3}{*}{\textbf{Model}}
		& \multicolumn{3}{c|}{\textbf{Automatic Evaluation}}
		& \multicolumn{3}{c}{\textbf{Human Evaluation}} \\
		& \multirow{2}{*}{\textbf{BLEU}}
		& \textbf{PARENT} % (Prec. / Rec. / F1)}
		& \multirow{2}{*}{\textbf{Avg Len.}}
		& \textbf{Faithful}
		& \textbf{Coverage}
		& \textbf{Fluency}\\
		&
		& (\textbf{Precision} / \textbf{Recall} / \textbf{F$_1$})
		&
		& \textbf{\%}
		& \textbf{\%}
		& \textbf{\%} \\
		\midrule
%		\color{gray} BERT-to-BERT-Wiki-Books~\citep{rothe2019leveraging} & \color{gray}  - & \color{gray} - \;/\; - \;/\; - & - & \color{gray}  77.20  & \color{gray}  4.38 & \color{gray}  99.20  \\
		BERT-to-BERT & \textbf{45.62} & 77.64 \;/\; 43.42 \;/\; 53.54 & 21.0 & 75.0  & 40.6 & \textbf{97.6} / 99.0 \\
	    Structure-Aware Seq2Seq & 45.36 & 73.98 \;/\; 44.02 \;/\; 52.81 & 23.1 & 66.4 & \textbf{40.6} & 89.0 / 99.6 \\
%		Pointer-Generator && & 87.2 & 4.23 & 99.0 \\
		Pointer-Generator, w/o lp & 41.07 & 77.59 \;/\; 42.12 \;/\; 52.10 & 19.1 & 79.6 & 39.4  & 93.6 / 96.2  \\
		BERT-to-LSTM, w/o lp & 42.50 & 77.11 \;/\; 40.62 \;/\; 50.94 & 20.1 & 75.6 & 38.4  & 95.2 / 98.6  \\
		\midrule
% 		This work &&& 88.2 & 3.98 & 93.3 \\
% 	    This work~(threshold=0.125) &&& 92.5 & 3.93 & 89.8 \\
%		Confident BERT-to-RNN & 33.30 & 77.98 \;/\; 37.21 \;/\; 47.90 & 16.6 & 85.2* & 3.90  & 92.3 / 94.1 \\
		Conf-PtGen, w/o lp, w/o null & 38.10 & 79.52 \;/\; 40.60 \;/\; 51.38 & 17.0 & \textbf{86.8*} & 37.8 & 94.6 / 95.8 \\
%	    \hspace{0.9 cm} +threshold=0.125 & 36.62 & \textbf{80.15} \;/\; 39.59 \;/\; 50.50 & 16.4 & \textbf{90.7*} & 4.01 & 91.6 / 92.2  \\
		Conf-T2LSTM, w/o lp, null 0.5 & 41.67 & 80.18 \;/\; 42.45 \;/\; 53.23 & 19.3 & 79.4 & 38.7 & 97.0 / 98.4 \\
		Conf-T2LSTM, w/o lp, null 0.8 & 40.21 & \textbf{80.38} \;/\; 41.47 \;/\; 52.41 & 18.7 & 84.8 & 38.7 & 89.6 / 92.0 \\
		Conf-T2LSTM, lp 2.0, null 0.8 & 44.21 & 78.83 \;/\; \textbf{44.11} \;/\; \textbf{54.35} & 21.4 & 80.8 & 40.2 & 85.4 / 91.0 \\
%		\midrule[.03em]
%		$^\dagger$This work~(\ours) & \textbf{37.3} & 18.5 & \textbf{34.7} \\
		\bottomrule
	\end{tabulary}
	\caption{Performance on WikiBio test set. The two fluency measures differ in whether to include sentences graded as \emph{Mostly Fluent}. Starred numbers are statistically significant against baselines ($p<.001$), by bootstrap test. The bottom block presents newly developed models in our work.}
	\label{tab:wikibio}
\end{table*}
%The top row (BERT-to-BERT-Wiki-Books) is faded since it was exposed to the test targets in pretraining.

%As typically done, the task is treated as sequence-to-sequence prediction; infoboxes are linearized, with field names and values separated by special tokens.

%\paragraph{Models}
We compare our method (the bottom two) against several strong baselines (the top four):
\begin{itemize}
\item \emph{BERT-to-BERT}~\citep{rothe2019leveraging}: A Transformer encoder-decoder model~\citep{vaswani2017attention} where the encoder and decoder are both initialized with BERT~\citep{devlin2018bert}. %The parameters between encoder and decoder are shared.

%BERT was trained on Wikipedia which overlaps with the test targets in the Wikibio dataset. Therefore, in addition to reporting this result, we also pre-trained a version of BERT on the Books corpus~\citep{zhu2015aligning} only, which we consider a more correct baseline. These two baselines are referred to as BERT-to-BERT-Wiki-Books and BERT-to-BERT-Books respectively.
    
\item \emph{Structure-aware Seq2Seq}~\citep{liu2017table}: A state-of-the-art method on WikiBio in terms of BLEU, which explicitly handles field names and table contents in an LSTM-based model.
    
\item \emph{Pointer-Generator}~\citep{see2017get}: Seq2Seq model with attention and copy mechanism.% (our implementation).

\item \emph{BERT-to-LSTM}: A Transformer encoder (initialized with BERT) to LSTM decoder model.% (our implementation).

\item \emph{Conf-PtGen} (Ours): A Pointer-Generator model with our proposed confident decoding.

\item \emph{Conf-T2LSTM} (Ours): A Transformer encoder to LSTM decoder model, with confident decoding.
\end{itemize}
%We built our confidence-oriented decoding method on the Pointer-Generator model using Tensorflow~\citep{abadi2016tensorflow} and employed the same hyper-parameter settings for both models. Our hyper-parameter search space and the chosen hyper-parameters are shown in Table~\ref{tab:hyperparameter}.In addition, we implemented a confidence-oriented BERT-to-RNN model, in which the encoder is a Transformer initialized with BERT checkpoint, and the decoder is a GRU~\citep{cho2014learning} with the same hidden size as the Transformer. The hyper-parameters are shown in Table~\ref{tab:hyperparameter}.
%Since our method is built on Pointer-Generator, we employed the same hyperparameter settings for both models.
%Both BERT-to-BERT and BERT-to-LSTM use the \textsc{bert-base-multilingual-uncased} checkpoint. For BERT-to-BERT, we use parameter sharing between the encoder and decoder, as it performs slightly better.
Here, the Pointer-Generator and BERT-to-LSTM are by our implementation, and serve as the base to our Conf-PtGen and Conf-T2LSTM models, respectively. More detailed experiment settings are found in the Appendix.

%For Pointer-Generator, we use GloVe~\citep{pennington2014glove} as the input word embedding and truncate the vocabulary size to 5,000. We use Tensorflow~\citep{abadi2016tensorflow} to build our systems.

%For both , we use Transformers with  12 layers, a hidden size of 768, filter size of 3072, and 12 attention heads. To initialize these models, we pre-trained an uncased BERT checkpoint on the Books corpus~\citep{zhu2015aligning} only, since the original BERT was trained on Wikipedia that overlaps with the test targets in WikiBio. Our pre-trained BERT trains positional embeddings for up to 512 positions and uses a vocabulary size of around 30k wordpieces~\citep{wu2016google}.

% Other hyper-parameter settings are given in Table~\ref{tab:hyperparameters}.

%For Pointer-Generator and Confident Pointer-Generator, we use GloVe~\citep{pennington2014glove} as the input word embedding and truncate the vocabulary size to 5,000; the two models share the same hyper-parameter settidfngs. For our confident decoding models, there are additional hyper-parameters $\rho$, $\gamma$, $K$ and $\lambda$ as defined in Section~\ref{sec:variational}. The hyper-parameters are shown in Table~\ref{tab:hyperparameters}.
%We use the Adam optimizer~\citep{DBLP:journals/corr/KingmaB14}.

\paragraph{Evaluation}
For automatic evaluation, we report BLEU~\citep{papineni2002bleu}, as well as PARENT~\citep{dhingra2019handling}, a metric that takes into account the data information, by aligning n-grams from the reference and prediction to the semi-structured input data, before computing their precision and recall. It is designed to mitigate the shortcomings of BLEU on  data-to-text generation.

For human evaluation, we obtain annotations on examples randomly chosen from predictions on the WikiBio test set, the same 500 for each model. Examples from different models are mixed and randomly shuffled, with model names hidden from the annotators.
%1000 randomly chosen examples for each model
%of model predictions on a set of 1000 randomly chosen examples on the test set.
We instruct the annotators to grade on each of 3 criteria: {\em faithfulness} (precision), {\em coverage} (recall), and {\em fluency}. Faithfulness assesses if all the information in the proposed sentence is supported by the table or the reference. A single hallucinated piece of information makes the sentence non-faithful. Coverage measures the number of table cells that contain information present in the sentence. Finally, fluency assesses if the sentence is clear, natural, and grammatically correct; raters choose among three options: \emph{Fluent} (clear, natural and grammatically correct; reads like a sentence found in a
book), \emph{Mostly Fluent} (with a few error, but mostly understandable), and \emph{Not Fluent} (with many errors and hardly understandable).

% \begin{itemize}
% \item Faithfulness (precision) - We define a sentence to be faithful if all the information in the proposed sentence is supported by the table or the reference. A single hallucinated piece of information makes the sentence non-faithful.
% \item  Coverage (recall) - The number of table cells that contain information present in the sentence.
% \item Fluency - A sentence is defined to be fluent if it is clear, natural, and grammatically correct. Raters choose among three options: \emph{Fluent}, \emph{Mostly Fluent}, \emph{Not Fluent}.
% \end{itemize}

An ideal system would always produce fluent and faithful text with high coverage. The output of our models and baselines, as well as the human evaluation data are publicly released.\footnote{The output of our models and baselines, with human evaluations, are available at
\url{https://drive.google.com/open?id=1Kg4hJkaK9gWCv7mxwBfHEQwAgF_TrwcE}. We will open-source our code as well.}

\paragraph{Results}
Table~\ref{tab:wikibio} shows the results. Despite achieving high BLEU scores, BERT-to-BERT and Structure-Aware Seq2Seq are less faithful according to human evaluation. Pointer-Generator is the most faithful among baselines, probably because its copy mechanism promotes verbatim copy from the source.
By applying our confident decoding method to the Pointer-Generator and BERT-to-LSTM respectively, we achieve clear improvement in faithfulness over the respective baselines.

Among the automatic metrics, PARENT precision and recall seem correlated to faithfulness and coverage respectively, and our approach achieves the highest precision and F1. BLEU, perhaps because of its length penalty that rewards longer generations, seems more correlated to coverage rather than faithfulness. Generally, it is easier for longer predictions to achieve higher coverage/recall, but harder to achieve faithfulness/precision.

In order to control recall while maintaining precision, we combine two techniques at inference time: The length penalty~\citep{wu2016google} which encourages longer generation, and the \textless{null}\textgreater ~token threshold (\S~\ref{sec:calibration}) which shuts out unconfident tokens. In Table~\ref{tab:wikibio}, Conf-PtGen does not use length penalty and is trained without \textless{null}\textgreater ~tokens (denoted ``w/o lp'' and ``w/o null'', respectively), so it tends to stop generation early when it is unconfident, which leads to shorter predictions and less coverage. In contrast, when Conf-T2LSTM incorporates \textless{null}\textgreater ~token with a moderate threshold (i.e.~null 0.5), it  improves \emph{both} precision and recall from the BERT-to-LSTM baseline, without sacrificing fluency. One can boost the precision and recall even more, by using length penalty to promote recall (e.g.~lp 2.0) and an aggressive \textless{null}\textgreater ~threshold (e.g.~null 0.8) to keep precision. This seems to cost some fluency, but most generations are still fluent (cf.~Appendix for generation examples).

%According to human evaluation, our approach gives a clear improvement in faithfulness over the baselines, with some drop in coverage. To further measure the validity of our confidence score, we postprocessed the output to remove words with lower confidence than $0.125$. This thresholding technique gives further gains to faithfulness, while sacrificing some fluency.

% Regarding the baselines, we see that BERT-to-BERT is the most fluent while Pointer Generator is the most faithful, suggesting that pretraining might help fluency while the copy mechanism can be valuable for faithfulness.

\paragraph{Ablation Test}
\label{sec:ablations}

\begin{table}[t!]
\footnotesize
	\centering
	\begin{tabulary}{0.5\textwidth}{l | c@{\hskip 0.4cm} c@{}}
		\toprule
	    \multirow{2}{*}{\textbf{Model}}
		& \multirow{2}{*}{\textbf{BLEU}}
		& \textbf{PARENT} \\ % (Prec. / Rec. / F1)}
		& & \textbf{(Precision / Recall / F$_1$)} \\
		\midrule
		Conf-PtGen & 38.10 & \textbf{79.52} \;/\; 40.60 \;/\; 51.38 \\
		\hspace{0.6 cm} -- Confidence & 39.39 & 78.77 \;/\; 41.55 \;/\; 52.08 \\
		\hspace{0.6 cm} -- Variational & \textbf{41.29} & 78.25 \;/\; \textbf{42.40} \;/\; \textbf{52.52} \\
		\hspace{0.6 cm} -- Calibration & 37.89 & 79.47 \;/\; 40.47 \;/\; 51.26 \\
		\midrule
		Pointer-Generator & 41.07 & 77.59 \;/\; 42.12 \;/\; 52.10 \\
		%\hspace{0.9 cm} Truncated & 35.50 & 77.68 \;/\; 38.16 \;/\; 48.66 \\
		\bottomrule
	\end{tabulary}
	\caption{Ablation tests on three components of Conf-PtGen.}
	\label{tab:ablation}
\end{table}

In this experiment, we assess the effects of three novel components in our confident decoding method:
%Our confident decoding method has three novel components:
\textbf{(1)} The design of a confidence score; \textbf{(2)} The variational Bayes objective with confident sub-sequence sampling; and \textbf{(3)} The calibration technique to re-rank output probabilities. We start from the Conf-PtGen, and in each test replace one component by a trivial alternative: \textbf{(1)} We compare with using the probability $P(y_t | \vy_{<t}, \vx)$ directly as confidence, and train models using the same hyper-parameters as Conf-PtGen. The results on the WikiBio test set are shown in Table~\ref{tab:ablation}, as ``-- Confidence''. \textbf{(2)} We compare with models trained by maximizing the ordinary log-likelihood, without sub-sequence sampling; the calibration technique is still applied (``-- Variational'').
\textbf{(3)} We disable the calibration technique (``-- Calibration'').

As we can see from Table~\ref{tab:ablation}, all three components improve PARENT precision. While the improvement by calibration is the smallest, the technique also improves PARENT recall and BLEU score at the same time, making it an easy choice. The other techniques trade recall for precision, making them useful for tasks that require a high degree of faithfulness. When all three components are disabled, the model is exactly the same as our implementation of the Pointer-Generator. Every component improves PARENT precision upon it as well. Especially, comparing Pointer-Generator with ``-- Variational'' shows again that calibration improves all metrics.

\paragraph{Sensitivity to Source}
\label{sec:srcsens}

\begin{figure}[t!]
\centering
\includegraphics[width=0.45\textwidth]{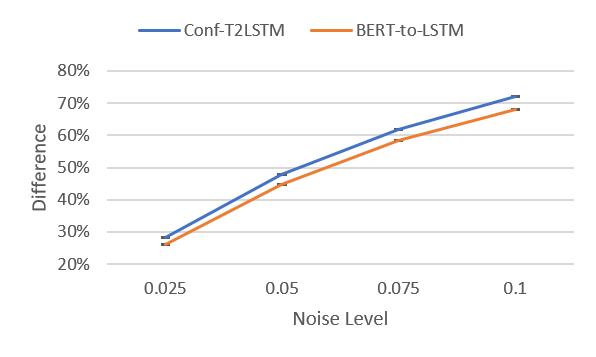}
\caption{More predictions changed for the Conf-T2LSTM model, when the source vectors are randomly set to $\mathbf{0}$ during decoding.}
\label{fig:srcsens}
\end{figure}

We have conjectured that making an encoder-decoder attend to the source whenever it is necessary can reduce hallucination; and we have clearly improved faithfulness by using a confident decoder that implements this conjecture. In this experiment, we show that Conf-T2LSTM is indeed more sensitive to the source than BERT-to-LSTM. The idea is to set all the source encoded vectors $\vs_1,...,\vs_{S} = \textbf{enc}(x_1,...,x_S)$ to $\mathbf{0}$ at some random steps during decoding, and see how many predictions changed. In Figure~\ref{fig:srcsens}, we show the results on the WikiBio validation set. As we increase the probability of source vectors to be set to $\mathbf{0}$, the predictions by Conf-T2LSTM changed more than BERT-to-LSTM. At each noise level, we decode $5$ times and plot the mean difference, as well as the standard deviation as error bar (almost indistinguishable from the lines in the chart). The exact predictions by both models are noise-sensitive: At a level of $0.1$, there are over $65\%$ of predictions changed already. However, most changes are subtle to human eyes; it is hard to glimpse any drop in generation quality.

\subsection{Results on WebNLG}
\label{sec:webnlgres}

\begin{table}[t!]
\footnotesize
	\centering
	\begin{tabulary}{0.5\textwidth}{l | c@{\hskip 0.3cm} c@{\hskip 0.2cm} c@{\hskip 0.2cm} c@{}}
		\toprule
	    \textbf{Model}
		& \textbf{BLEU}
		& \textbf{NIST} & \textbf{METEOR} & $\textbf{ROUGE}_L$ \\
		\midrule
		T2LSTM-att & 65.35 & 11.14 & \textbf{0.4615} & 0.7648 \\
		T2LSTM & 65.15 & 11.15 & \textbf{0.4615} & \textbf{0.7685} \\
		Conf-T2LSTM & \textbf{65.58} & \textbf{11.21} & 0.4601 & 0.7676 \\
		\midrule
		OpenNMT & 48 & 8.09 & 0.36 & 0.65 \\
		%\hspace{0.9 cm} Truncated & 35.50 & 77.68 \;/\; 38.16 \;/\; 48.66 \\
		\bottomrule
	\end{tabulary}
	\caption{Results on WebNLG v2.1 constrained split~\citep{shimorina2018handling}.}
	\label{tab:webnlg}
	%\vspace{-0.3cm}
\end{table}

The WebNLG dataset has more controlled data format and generation patterns than WikiBio, making it a suitable benchmark for data-to-text models. Although the issue of hallucination is not severe on this dataset, we use it to compare modifications we made on the encoder-decoder architecture vs.~the conventional designs.
In particular, we compare: (i) \emph{T2LSTM-att}, a 12-layer Transformer encoder to LSTM decoder architecture, with conventional attention mechanism; (ii) \emph{T2LSTM}, with our modified attention as defined in Eq.~\eqref{eq:attention2}\eqref{eq:rnn2}; (iii) \emph{Conf-T2LSTM}, with our confident decoding. All three models use the sentence-piece tokenizer~\citep{kudo2018sentencepiece} with a vocabulary size of $4,000$, and length penalty $1.0$ at inference. For Conf-T2LSTM, the \textless{null}\textgreater ~threshold is set to $0.5$. We evaluate using BLEU~\citep{papineni2002bleu}, NIST~\citep{doddington2002automatic}, METEOR~\citep{banerjee2005meteor},
and $\text{ROUGE}_L$~\citep{lin2004rouge} computed using the evaluation scripts for the E2E Challenge~\citep{duvsek2018findings}.

According to results shown in Table~\ref{tab:webnlg}, our modeling enhancements do not degrade performance in the presence of clean references. Compared to an OpenNMT~\citep{klein2017opennmt} baseline with the best delexicalisation and copying setting reported by~\citet{shimorina2018handling}, our models demonstrate strong performance. We do not include other baselines such  as~\citet{ferreira2019neural} and~\citet{kale2020text} because they report numbers on an older version of the WebNLG corpus.

\section{Discussion and Related Work}
\label{sec:related}

Many ideas have been explored in text generation to achieve more accurate predictions, such as
%our work aligns with
%Broaderly, more accurate text generation has been achieved by
learning neural templates~\citep{wiseman2018learning}, separating content selection from generation~\citep{zhou-etal-2017-selective,gehrmann2018bottom,puduppully2019data},
MaskGAN~\citep{fedus2018maskgan},
%constrained vocabulary decoding~\citep{wu-aaai18},
%reinforcement learning-based  rewards~\citep{paulus2018deep,pasunuru2018multi},
entity modeling~\citep{puduppully2019entity},
data augmentation~\citep{ma-etal-2019-key,kedzie2019good}, \emph{etc}. Among them, improving the faithfulness
%Improving the fidelity of text generation
is an emerging research topic that has been tackled by a variety of works recently. Concurrent to our work, \citet{matsumaru2020improving} empirically found that removing unfaithful instances from the training data can reduce hallucination in headline generation, while \citet{kang2020improved} proposed a loss truncation training framework that can remove such noise in a principled manner. However, removing entire examples from the loss is not practical for noisy datasets such as Wikibio, where 62\% of examples exhibit divergence. \citet{wang2020towards} and \citet{shen2020neural} tackle this problem by adding additional terms to the loss that enforce alignment between source and target, while the ``goodness'' of such alignment relies on heuristics specific to the task and data. In contrast, the prior knowledge we exploit in this work is more general, as it does not depend on any source-specific data structure. Our method has the potential to be adapted to other generation tasks such as document summarization and machine translation, or combined with other approaches. \citet{li2020posterior} also propose to control neural text generation by posterior regularization, but they still rely on heuristics such as surface matching between source and target. \citet{harkous2020text} address the problem by decoder re-ranking, while the ranker is trained on heuristically extracted faithful data-text pairs. Complementary to all these works, our approach develops a deeper understanding of the encoder-decoder architecture itself, by carefully reconsidering its attention and language modeling components.

\bibliography{confidence_decoding}
\bibliographystyle{acl_natbib}

\newpage
\appendix
\onecolumn

\section{Generation Examples and Analysis of Hallucination}
\label{sec:genexamples}

In the following, we show some typical cases where the Pointer-Generator baseline hallucinates but the Conf-PtGen does not. In most of the cases, the Conf-T2LSTM models do not hallucinate either. Hallucinated parts are colored \textcolor{orangered}{red}.
The examples are taken from the WikiBio validation set.

In the first six examples (i.e.~``\textit{Frank Lino}'', ``\textit{Rohan Robertson}'', ``\textit{Walter Smallwood}'', ``\textit{Nellie Wong}'', ``\textit{Hal Bedsole}'' and ``\textit{Constant Vanden Stock}''), some information is missing in the table while the Pointer-Generator baseline made it up. Our confident decoder models learn to omit the missing fields, although by doing this, some of the generated sentences become not fluent.

In the next example (i.e.~``\textit{Richard Lloyd}''), the baseline seems to have learned weird language modeling from some similar training points, and tries to generate more than the table contents; our confident decoder models generate correctly.

In the last two examples (i.e.~``\textit{Robert I. Marshall}'' and ``\textit{Thomas Edwards}''), there are corresponding fields in the table but the Pointer-Generator baseline didn't learn to generate correctly, possibly because these fields should not be simply copied. Our confident models generate more faithfully to the source.

\hrulefill

\begin{description}
\item[Table Info.] The occupation is missing in the table.
\item[Reference] \textit{Frank ``Curly'' Lino (born October 30, 1938 Brooklyn)  is a Sicilian-American Caporegime in the Bonanno crime family who later became an informant.}
\item[Pointer-Generator] \textit{Frank Lino (born October 30, 1938 in Brooklyn, New York, United States) is an American \textcolor{orangered}{criminal defense attorney}.}
\item[Conf-PtGen] \textit{Frank Lino (born October 30, 1938 in Brooklyn, New York, United States) is an American.}
\item[Conf-T2LSTM, w/o lp, null 0.5] \textit{Frank Lino (born October 30, 1938) is an American \textcolor{orangered}{actor}.}
\item[Conf-T2LSTM, w/o lp, null 0.8] \textit{Frank Lino (born October 30, 1938 Gravesend , Brooklyn , New York) is a former American}
\item[Conf-T2LSTM, lp 2.0, null 0.8] \textit{Frank Lino (born October 30, 1938 Gravesend, Brooklyn, New York) is a former American}
\end{description}

\hrulefill

\begin{description}
\item[Table Info.] The team name is missing in the table.
\item[Reference] \textit{Rohan Robertson (born 21 August 1961) is a former Australian rules footballer who played for North Melbourne in the Victorian Football League (VFL) between 1985 and 1988.}
\item[Pointer-Generator] \textit{Rohan Robertson (born 21 August 1961) is a former Australian rules footballer who played \textcolor{orangered}{with Carlton} in the Victorian Football League.}
\item[Conf-PtGen] \textit{Rohan Robertson (born 21 August 1961) is a former Australian rules footballer who played in the Victorian Football League.}
\item[Conf-T2LSTM, w/o lp, null 0.5] \textit{Rohan Robertson (born 21 August 1961) is a former Australian rules footballer who played \textcolor{orangered}{with Carlton} in the Victorian Football League (VFL).}
\item[Conf-T2LSTM, w/o lp, null 0.8] \textit{Rohan Robertson (born 21 August 1961) is a former Australian rules footballer who played \textcolor{orangered}{with Carlton} in the Victorian Football League (VFL).}
\item[Conf-T2LSTM, lp 2.0, null 0.8] \textit{Rohan Robertson (born 21 August 1961) is a former Australian rules footballer who played \textcolor{orangered}{with Carlton} in the Victorian Football League (VFL).}
\end{description}

\hrulefill

\begin{description}
\item[Table Info.] The middle name is missing in the table.
\item[Reference] \textit{Walter Clayton Smallwood (April 24, 1893 -- April 29, 1967) was a professional baseball pitcher from 1913 to 1931.}
\item[Pointer-Generator] \textit{Walter \textcolor{orangered}{Herbert} Smallwood (April 24, 1893 -- April 29, 1967) was a pitcher in major League Baseball.}
\item[Conf-PtGen] \textit{Walter Smallwood (April 24, 1893 -- April 29, 1967) was a pitcher in major League Baseball.}
\item[Conf-T2LSTM, w/o lp, null 0.5] \textit{Walter \textcolor{orangered}{Henry} Smallwood (April 24, 1893 -- April 29, 1967) was a major league baseball pitcher.}
\item[Conf-T2LSTM, w/o lp, null 0.8] \textit{Walter Smallwood (April 24, 1893 -- April 29, 1967) was a major league baseball pitcher.}
\item[Conf-T2LSTM, lp 2.0, null 0.8] \textit{Walter Smallwood (April 24, 1893 -- April 29, 1967) was a major league baseball pitcher.}
\end{description}

\hrulefill

\begin{description}
\item[Table Info.] The residence address is not stated in the table.
\item[Reference] \textit{Nellie Wong (born 12 September 1934) is a poet and activist for feminist and socialist causes.}
\item[Pointer-Generator] \textit{Nellie Wong (born September 12, 1934 in Oakland, California) is an American poet, activist, \textcolor{orangered}{feminist, and feminist activist who lives and works in Los Angeles}, California, United States, where she}
\item[Conf-PtGen] \textit{Nellie Wong (born September 12, 1934) is an American poet and activist.}
\item[Conf-T2LSTM, w/o lp, null 0.5] \textit{Nellie Wong (born September 12, 1934) is an American poet and activist.}
\item[Conf-T2LSTM, w/o lp, null 0.8] \textit{Nellie Wong (born September 12, 1934) is an American poet}
\item[Conf-T2LSTM, lp 2.0, null 0.8] \textit{Nellie Wong (born September 12, 1934) is an American poet}
\end{description}

\hrulefill

\begin{description}
\item[Table Info.] The name field is missing in the table.
\item[Reference] \textit{Harold Jay ``Hal'' Bedsole (born December 21, 1941) is a retired American football player.}
\item[Pointer-Generator] \textit{\textcolor{orangered}{Gene} (born December 21, 1941) is a former American football tight end in the National Football League.}
\item[Conf-PtGen] \textit{(Born December 21, 1941) is a former American Football tight end in the National Football League for the Minnesota Vikings.}
\item[Conf-T2LSTM, w/o lp, null 0.5] \textit{Hal Bedsole (born December 21, 1941) is a former American football tight end in the national football league.}
\item[Conf-T2LSTM, w/o lp, null 0.8] \textit{Hal Bedsole (born December 21 , 1941) is a former american football tight end in the national football league.}
\item[Conf-T2LSTM, lp 2.0, null 0.8] \textit{Hal Bedsole (born December 21, 1941 Chicago, Illinois) is a former American football tight end in the national football league for the Minnesota Vikings and Minnesota Vikings.}
\end{description}

\hrulefill

\begin{description}
\item[Table Info.] The nationality and occupation is missing in the table.
\item[Reference] \textit{Constant Vanden Stock (; 13 june 1914 -- 19 April 2008) was the honorary president and former president and player of Belgian football club R.S.C. Anderlecht.}
\item[Pointer-Generator] \textit{Constant Vanden Stock (June 13, 1914 -- April 19, 2008) was an \textcolor{orangered}{American Figure Skater}.}
\item[Conf-PtGen] \textit{Constant Vanden Stock (June 13, 1914 -- April 19, 2008) was a}
\item[Conf-T2LSTM, w/o lp, null 0.5] \textit{Constant Vanden Stock (June 13, 1914 -- April 19, 2008) was an \textcolor{orangered}{American politician}.}
\item[Conf-T2LSTM, w/o lp, null 0.8] \textit{Constant Vanden Stock (June 13, 1914 -- April 19, 2008) was an \textcolor{orangered}{American}}
\item[Conf-T2LSTM, lp 2.0, null 0.8] \textit{Constant Vanden Stock (June 13, 1914 -- April 19, 2008) was an \textcolor{orangered}{American}}
\end{description}

\hrulefill

\begin{description}
\item[Table Info.] Available fields in the table are name, nationality, years and teams.
\item[Reference] \textit{Richard Lloyd (18 February 1945 -- 30 March 2008) was a British racing car driver and founder of multiple sports car and touring car teams.}
\item[Pointer-Generator] \textit{Richard Lloyd Lloyd is a British Racing driver who won the GTI Engineering \textcolor{orangered}{Championship in 1982}, driving with GTI Engineering, Richard Lloyd Racing of GTI Engineering, and Lloyd Lloyd Racing \textcolor{orangered}{at the age of 14}.}
\item[Conf-PtGen] \textit{Richard Lloyd is a former racing driver.}
\item[Conf-T2LSTM, w/o lp, null 0.5] \textit{Richard Lloyd was a British racing driver.}
\item[Conf-T2LSTM, w/o lp, null 0.8] \textit{Richard Lloyd was a British racing driver.}
\item[Conf-T2LSTM, lp 2.0, null 0.8] \textit{Richard Lloyd was a British racing driver.}
\end{description}

\hrulefill

\begin{description}
\item[Table Info.] The district field has value ``3rd''.
\item[Reference] \textit{Robert I. Marshall (born October 16, 1946 in Wilmington, Delaware) is an American politician and a democratic member of the Delaware Senate since January 9, 1979 representing district 3.}
\item[Pointer-Generator] \textit{Robert I. Marshall (born October 16, 1946 in Wilmington, Delaware) is an American politician and a Democratic member of the Delaware Senate since January 9, 1979 representing district \textcolor{orangered}{41}.}
\item[Conf-PtGen] \textit{Robert I. Marshall (born October 16, 1946 in Wilmington, Delaware) is an American politician and a Democratic member of the Delaware Senate since January 9, 1979.}
\item[Conf-T2LSTM, w/o lp, null 0.5] \textit{Robert I. Marshall (born October 16 , 1946 in Wilmington, Delaware) is an American politician and a Democratic member of the Delaware Senate since January 9, 1979 representing district 3 .}
\item[Conf-T2LSTM, w/o lp, null 0.8] \textit{Robert I. Marshall (born October 16, 1946 in Wilmington, Delaware) is an American politician and a Democratic member of the Delaware Senate since January 9, 1979 representing district 3.}
\item[Conf-T2LSTM, lp 2.0, null 0.8] \textit{Robert I. Marshall (born October 16, 1946 in Wilmington, Delaware) is an American politician and a Democratic member of the Delaware Senate since January 9, 1979 representing district 3.}
\end{description}

\hrulefill

\begin{description}
\item[Table Info.] The table has a field ``known for'': English and Welsh dictionary.
\item[Reference] \textit{Thomas Edwards (Caerfallwch), (1779 -- 1858), was a Welsh author.}
\item[Pointer-Generator] \textit{Thomas Edwards (1779 -- 4 June 1858) was an English author, UNK, and UNK, who \textcolor{orangered}{spent most of his life in the English and English literature of the English} and English dictionary literature.}
\item[Conf-PtGen] \textit{Thomas Edwards (1779 -- 4 June 1858) was a Welsh author.}
\item[Conf-T2LSTM, w/o lp, null 0.5] \textit{Thomas Edwards (1779 -- 4 June 1858) was a Welsh author.}
\item[Conf-T2LSTM, w/o lp, null 0.8] \textit{Thomas Edwards (1779 -- 4 June 1858) was a Welsh and Welsh dictionary}
\item[Conf-T2LSTM, lp 2.0, null 0.8] \textit{Thomas Edwards (1779 -- 4 June 1858) was a Welsh and Welsh dictionary.}
\end{description}

\hrulefill

\section{Interplay between Attention Score, Base Language Model, and Confidence Score}
\label{sec:interplay}

Figure~2 in our main paper showed an example of the learned attention score, base language model probability and confidence score of our Conf-T2LSTM model. The example is taken from the WikiBio validation set, using the reference sentence.

\begin{figure}[b]
\centering
\caption{More examples of the learned attention score, base language model probability and confidence score.}
\label{fig:appendix_scores}
\includegraphics[scale=0.38,viewport=0 0 42cm 13cm,clip]{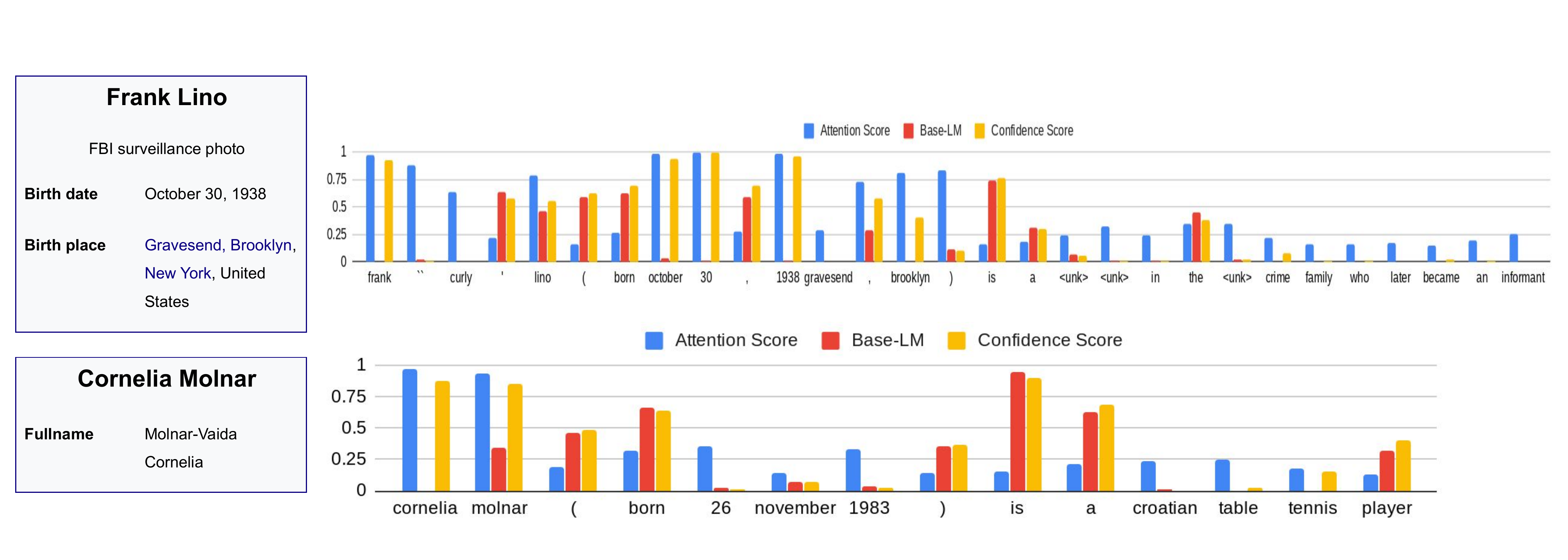}
\end{figure}

In order to further illustrate the mechanism of our approach, in Figure~\ref{fig:appendix_scores} we show more examples of the scores, learned by our Conf-PtGen model. Compared to Conf-T2LSTM, Conf-PtGen has a copy mechanism, and the attention scores seem more sensitive to missing fields in the table. In Figure~\ref{fig:appendix_scores}, \textit{Occupation} is missing in the \textit{Frank Lino} table, and the \textit{Cornelia Molnar} table only has a name. Our model successfully detected the tokens not supported by the table.

\section{Human Evaluation Instructions}

We show the detailed instructions for our human evaluation in the following.
We have discussed with the lead annotators about many other examples as well. We have made sure that: (a) Valid inferences (e.g.~inferring nationality from birth place) are considered faithful; (b) If a piece of information exists or can be inferred from the table, the corresponding cell should be highlighted, even if the information was also in the background knowledge; (c) Only one cell should be highlighted for one piece of information.

\hrulefill

\textit{A writer has the following background knowledge and is given a table:}\\
\textless{Reference sentence shown as background knowledge}\textgreater\\
\textless{Table}\textgreater

\textit{The writer read the table and produced the following sentence:}\\
\textless{Sentence generated by model}\textgreater\\
\textit{We wish to evaluate the quality of the sentence.}

1. How \textbf{fluent} is the sentence?
\begin{itemize}
\item Fluent: It is clear, natural, and the grammar is correct. It reads like if it was found in a book.
\item Mostly Fluent: It has a few errors or it does not sound natural, but you can understand it.
\item Not Fluent: It has many errors and/or you can hardly understand it.
\end{itemize}

Examples:

\begin{table}[h!]
	\centering
\begin{tabulary}{0.3\textwidth}{p{6.5cm} | p{6.5cm}}
	\toprule
	Sentence & Decision \\
	\midrule
	alfred angas scott ( 1875 - 1923 ) was a motorcycle designer born in manningham bradford , who lived in
	& \textbf{Not fluent}: the sentence stops abruptly. It is unnatural, it does not make any sense. \\
	\midrule
	alfred angas scott ( 1875 - 1923 ) was a motorcycle designer designer born in manningham bradford .
	& \textbf{Mostly fluent}: the repetition ``designer designer'' is not natural, but the sentence makes sense. \\
	\midrule
	alfred angas scott ( 1875 - 1923 ) was a motorcycle designer , was born in manningham bradford , lived in england u.k. , was british by nationality , and was buried at the undercliffe cemetery in bradford .
	& \textbf{Mostly fluent}: there are no mistakes, but the sentence contains too much information. This is unnatural. \\
	\bottomrule
\end{tabulary}
	%\caption{ADD HERE.}
\end{table}

2. Please compare carefully the content of the sentence to the content of the table. How many cells from the table did the writer use to produce the sentence? (Click on the cells in the table above to update the counter)

3. A sentence is \textbf{faithful} if it contains only information supported by the table or the writer's background knowledge. It should not add any additional information, even if the information is true or interesting. Please compare once again the content of the sentence to the content of the table and background knowledge. How faithful is the sentence?
\begin{itemize}
\item Faithful: every part of the sentence is supported by the table and/or background knowledge.
\item Mostly Faithful: every part of the sentence can be linked to some evidence in the table or the background knowledge, but it is not fully supported. This should only be used for rare edge cases.
\item Not Faithful: The sentence contains information that is not supported by the table or background knowledge.
\end{itemize}
The examples are based on the following background knowledge and table:

alfred angas scott ( 1875 - 1923 ) was a british motorcycle designer .

\begin{table}[h!]
	\centering
\begin{tabulary}{0.3\textwidth}{l p{2.4cm} l l l p{2.8cm}}
	\toprule
	birth date & birth place & death date & nationality & residence & occupation \\
	\midrule
    1875 & manningham, bradford, united kingdom & 1923 & british & england, uk. & motorcycle designer and manufacturer \\
	\bottomrule
\end{tabulary}
	%\caption{ADD HERE.}
\end{table}

\newpage

Examples:

\begin{table}[h!]
	\centering
\begin{tabulary}{0.3\textwidth}{p{6.5cm} | p{6.5cm}}
	\toprule
	Sentence & Decision \\
	\midrule
	alfred angas scott ( 1875 - 1923 ) was a british motorcycle designer and founder of the scott motorcycle company .
	& \textbf{Coverage}: 4 \newline
	\textbf{Not faithful}: the table does not mention that A.~A.~Scott was the founder of the Scott Company .\\
	\midrule
	alfred angas scott ( 1875 - 1923 ) was a european motorcycle designer .
	& \textbf{Coverage}: 4 \newline
	\textbf{Faithful}: All the information is supported because England is located in Europe.\\
	\bottomrule
\end{tabulary}
	%\caption{ADD HERE.}
\end{table}

\section{Detailed Experiment Settings}

In our experiments, we use Transformers of 12 layers, a hidden size of 768, filter size of 3072, and 12 attention heads. The LSTM has a hidden size of 256 and memory size of 1024.
Both the BERT-to-BERT and BERT-to-LSTM models use the \textsc{bert-base-multilingual-uncased} checkpoint, with a vocabulary size of 105k.
For BERT-to-BERT, we use parameter sharing between the encoder and decoder, as it performs slightly better.
For Pointer-Generator, we use GloVe~\citep{pennington2014glove} as the input word embedding and truncate the vocabulary size to 5,000. We use Tensorflow~\citep{abadi2016tensorflow} to build our systems.

For training, we use the Adam optimizer~\citep{DBLP:journals/corr/KingmaB14}, and the learning rate is set to $0.00005$ for BERT-to-LSTM, and $0.0005$ for Pointer-Generator. We use early-stopping based on validation loss to determine the training epochs.
The dropout rates in the LSTM model are especially important for appropriate training; we use dropout rate $0.5$ for the input layer of LSTM, and RNN-dropout $0.2$ for the memory; the dropout rate applied to the attention layer is $0.2$ for WikiBio and $0.1$ for WebNLG. The number $K$ of samples we use for the Monte Carlo method of our variation Bayes loss is set to $8$. For decoding, we use a beam size of $8$.

\end{document}